\documentclass[11pt]{article}

\usepackage{acl}
\usepackage{times}
\usepackage{latexsym}

\usepackage[T1]{fontenc}
\usepackage[utf8]{inputenc}

\usepackage{microtype}

\usepackage{inconsolata}

\usepackage{graphicx}

\title{From Symbolic to Natural-Language Relations: Rethinking Knowledge Graph Construction in the Era of Large Language Models}

\author{Kanyao Han, Yushang Lai\\
Walmart Global Tech\\
\texttt{\{kanyao.han, yushang.lai\}@walmart.com}}

\begin{document}
\pagestyle{plain}
\thispagestyle{plain}

\maketitle
\begin{abstract}
Knowledge graphs (KGs) have commonly been constructed using predefined symbolic relation schemas, typically implemented as categorical relation labels. This design has notable shortcomings: real-world relations are often contextual, nuanced, and sometimes uncertain, and compressing it into discrete relation labels abstracts away critical semantic detail. Nevertheless, symbolic-relation KGs remain widely used because they have been operationally effective and broadly compatible with pre-LLM downstream models and algorithms, in which KG knowledge could be retrieved or encoded into quantified features and embeddings at scale. The emergence of LLMs has reshaped how knowledge is created and consumed. LLMs support scalable synthesis of domain facts directly in concise natural language, and prompting-based inference favors context-rich free-form text over quantified representations. This position paper argues that these changes call for rethinking the representation of relations themselves rather than merely using LLMs to populate conventional schemas more efficiently. We therefore advocate moving from symbolic to natural-language relation descriptions, and we propose hybrid design principles that preserve a minimal structural backbone while enabling more flexible and context-sensitive relational representations.
\end{abstract}

\section{Introduction}

Knowledge graphs (KGs) are widely used in domain- and task-specific applications to compensate for the lack of specialized knowledge in pre-trained models and other algorithmic systems. When training data or model/algorithm capacity alone is insufficient to capture domain-specific facts and relationships, KGs offer an explicit, structured representation of external knowledge that supports inference and improves performance on knowledge-intensive tasks \citep{fensel2020knowledge,hogan2021knowledge,qu2022review}. A KG is typically represented as a collection of relational triples of the form $[Entity_1, Relation, Entity_2]$, optionally augmented with attributes associated with entities and relations. Entities correspond to real-world concepts (e.g., abstract notions) or instances (e.g., persons, organizations, and events), while relations represent connections between pairs of entities. This structured representation enables KGs to serve as an explicit, machine-recognizable abstraction of domain knowledge that can be integrated with downstream models and algorithms.

Central to the KG paradigm is the design and expressiveness of relations, as they determine what types of knowledge can be represented. The most widely adopted approach to KG construction relies on defining a set of symbolic relation categories with a well-defined schema that specifies how entities should be connected \citep{hogan2021knowledge,radevski2023linking}, such as synonymy, subclass, friend, affiliation, and many others \citep{fellbaum2010wordnet,vrandevcic2014wikidata,steiner2012adding}.

Although symbolic relation categories capture only simplified abstractions of real-world relationships, they have historically provided a practical and effective paradigm for knowledge extraction and representation \citep{bordes2013translating,trouillon2016complex,xia2019random}. In this framework, relation extraction \citep{pawar2017relation} in KG construction is typically formulated as a classification problem: given a pair of entities co-occurring in text(s) or graph(s), models are used to determine whether a relation exists between them and, if so, to assign it to one of a set of relation types. This formulation aligns well with earlier models that take text/graph features or embeddings as input and symbolic relation labels as output, enabling efficient and accurate construction of KGs from large-scale text corpora or extend existing KGs. The resulting KGs with symbolic relation categories can be readily used as factual information or encoded into quantified features, embeddings, or graph structures, making them compatible with a wide range of downstream tasks such as question answering, recommendation, clustering, and indexing \citep{sun2018recurrent,xiao2019knowledge,dai2020survey,yani2021challenges,ren2023fact,opsahl2024fact}.

However, this paradigm also introduces inherent limitations. Since relation types are usually over-simplified, predefined, and discrete \citep{yates2007textrunner,niklaus2018survey,pai2024survey}, the resulting representations tend to abstract away much of the contextual and situational information present in real-world relations and thus restrict the ability to capture nuanced, context-dependent relations \citep{emirbayer1994network,boccaletti2014structure}. Consequently, many aspects of real-world relationships are either simplified or omitted altogether, revealing a mismatch between the inflexible structure of symbolic relation schemas and labels and the complexity of naturally occurring relational knowledge.

Recent advances in large language models (LLMs) have fundamentally changed the landscape of both knowledge resource construction and their downstream applications. For resource creation \citep{ding2024data,tan2024large, yang2025comprehensive}, LLMs enable scalable extraction and synthesis of information from massive data, making it feasible to represent knowledge using concise natural-language descriptions rather than numeric or categorical labels. For downstream usage \citep{baek2023knowledge,dong2024survey,shu2024knowledge,wang2024mgsa}, prompting-based inference naturally favors nuanced, context-rich natural language, as such models are designed to interpret and reason over free-form text rather than discrete categorical relation types. As a result, relations between pairs of entities in KG expressed in natural language can be more directly consumed and exploited by LLM-based downstream applications than those represented as traditional symbolic relations.

While recent studies in KG, ontology, or taxonomy construction have increasingly leveraged LLMs, most existing efforts are still within the traditional symbolic paradigm — using LLMs primarily as tools to populate or refine schemas or help symbolic relation prediction \citep{wei2023kicgpt,fathallah2024neon,liang2024survey,zhu2024llms}. In contrast, this position paper argues for a more fundamental paradigm shift: Rather than merely employing LLMs to construct conventional KGs with symbolic relations more efficiently, we advocate rethinking the representation of relations themselves. Specifically, we propose moving from symbolic relation categories toward natural-language-based relational representations that better align with the capacity of LLMs. Such a shift enables richer, more flexible, and more context-sensitive representations of knowledge, and is well aligned with the way LLMs are increasingly used in practice.

In this position paper, we first provide a brief review of earlier KG practices and explain why symbolic relations became dominant due to their scalability and compatibility with prior downstream models. We then contrast this tradition with the capabilities and preferences of LLM-centered workflows, highlighting the gap that emerges when complex relational knowledge must be compressed into discrete symbolic labels. Building on this analysis, we articulate conceptual and operational arguments for a paradigm shift from symbolic to natural-language relation representation, and propose a set of principles for hybrid KG designs that preserve structural usability while enabling contextual richness. Finally, we outline concrete research directions for KG construction, refinement, traversal, embedding, and evaluation under this new paradigm.

% \begin{itemize}
%     \item Reviewing existing KG construction practices: We analyze how current knowledge graph construction practices increasingly incorporate LLMs for tasks such as entity and relation extraction, while still largely adhering to traditional symbolic relation schemas.
    
%     \item Examining limitations of symbolic and OpenIE-style relations: We discuss the constraints of symbolic relations and simple OpenIE-style representations, particularly their limited ability to capture contextual, nuanced, and semantically rich relationships that frequently arise in downstream LLM-based applications.

%     \item Exploring natural-language relations as an alternative: We investigate the strengths of natural-language-based relational representations, which better align with how LLMs encode, reason over, and utilize knowledge in practice.

%     \item Addressing challenges and positioning natural-language relations: We analyze key challenges in adopting natural-language relations, including normalization, evaluation, and operational integration, and argue that such relations should be viewed as a complementary extension—rather than a replacement—to symbolic relations, which remain valuable in many application settings.
% \end{itemize}

\section{KGs in Prior Literature}
\subsection{Forms of Relation Representation}

From the perspective of relation representation, existing KGs can be broadly categorized into two forms: symbolic relation and OpenIE relation.

\textbf{Symbolic Relation KGs:} Symblic relations come from a predefined schema consisting of discrete relation categories. Each edge in the graph corresponds to a symbolic relation label with well-defined semantics \citep{fensel2020knowledge,unni2022biolink,zhou2022schere}. Common examples include linguistic relations (e.g., synonymy-of) \citep{fellbaum2010wordnet}, taxonomic relations (e.g., child-of/subclass-of) \citep{vrandevcic2014wikidata,han2020wikicssh}, social or interpersonal relations (e.g., friend-of) \citep{cao2020building}, affiliation or locational relations (e.g., works-at, lives-in) \citep{steiner2012adding}, event-centered relations (e.g., participated-in, occurred-at) \citep{guan2022event}, and domain-specific semantic relations (e.g. paper-studies-topic) \citep{zhu2025context}.

\textbf{Open Information Extraction (OpenIE) KGs:} OpenIE approaches have also been slightly explored for the construction of KGs in prior studies \citep{martinez2018openie}, which represent relations using surface linguistic expressions (e.g., ``grew up in'', ``worked for'') , typically verbs or verb phrases extracted directly from text \citep{radevski2023linking}. However, OpenIE KGs are less used in downstream tasks as it contains lots of noise and non-canonical relations that are not compatible with domain- or task-specific applications \citep{romero2023mapping}. Canonicalizing the open-ended surface linguistic relations into a fixed, normalized set of symbolic labels is often required for domain- or task-specific applications, but has been hardly addressed so far \cite{niklaus2018survey}.

\subsection{Downstream Usages}

The usages of KGs in downstream tasks can be broadly categorized into four groups: (1) factual information retrieval, (2) representation learning where KGs serve as structural supervision for embedding-based models, (3) organization, indexing, and navigation of knowledge artifacts, and (4) new knowledge discovery through link prediction and KG completion.

\textbf{Factual Information Retrieval:} In this usage, KGs function as structured stores of factual information, from which entities, relations, attributes, or subgraphs are retrieved as needed through entity matching or graph traversal. Retrieved information may be presented as answers in fact-based question answering \citep{bao2016constraint,yani2021challenges}, used as auxiliary evidence to augment downstream models \citep{opsahl2024fact,zhu2025knowledge}, or converted as additional data for model fine-tuning or training \citep{tian2024kg}. 

\textbf{Representation Learning:} KGs are used primarily as structural supervision to shape learned representations, most commonly entity embeddings \citep{dai2020survey,he2020knowledge,ge2024knowledge}. Relation structure influences how representations are learned during training, but the explicit graph is typically not consulted at inference time. Downstream tasks such as recommendation \citep{grad2017graph,sun2018recurrent,shokrzadeh2024knowledge}, ranking \citep{xiong2017explicit,ren2023fact}, and clustering \citep{xiao2019knowledge,gad2020excut} are often augmented by KG embeddings.

\textbf{Organization, Indexing, and Navigation of Knowledge Artifacts:} In this usage, the KG is employed as structural backbones for organizing and indexing information resources, such as documents, texts, multimedia, or archival records. It supports efficient indexing, faceted search, browsing, and navigation by providing controlled vocabularies and explicit conceptual relationships, enabling faster search/retrieval and more coherent organization of large knowledge collections \citep{wenige2020automatic,limani2021scholarly}.

\textbf{New Knowledge Discovery via New Link Prediction:} In this paradigm, the KG serves as a basis for discovering new or missing links between unlinked entities. Models estimate the plausibility of unobserved links between entities, producing candidate relations or triples that extend the existing graph \citep{zamini2022review,cai2023temporal,liang2025exploring}. This usage treats the KG not only as a record of known facts, but as a basis for hypothesizing additional knowledge under the assumptions imposed by the graph’s structure and schema.

\subsection{Limitations of Existing KGs}

While pre-defined relation schemas and the graph structure enable scalability and broad computational compatibility with traditional models and systems as well as LLMs, prior literature in computer science, biomedical science, and social science have long recognized the limitations of the mainstream KGs with symbolic relations from different perspectives.

\textbf{Rigidity of Predefined Symbolic Relation Schemas:} Prior computer science research has discussed the constraints from static and predefined relation, which determine in advance what kinds of facts can be represented, retrieved, organized, or inferred \citep{yates2007textrunner,niklaus2018survey,pai2024survey}. As a result, downstream usages are all bounded by the scope of the relation schema. Relational dimensions that fall outside the predefined categories are systematically omitted, and extending coverage typically requires costly schema redesign and data reprocessing. This rigidity also limits generalization: relations not explicitly anticipated by the schema cannot be easily represented, even when they are salient or frequent in real-world data. 

\textbf{Lack of Canonicalization in OpenIE KGs:} OpenIE KGs are the most discussed alternatives to relax the rigidity of symbolic relation schemas by expressing relations as open-ended verb or verb-phrase predicates extracted from text, enabling broader coverage without predefined ontologies \citep{martinez2018openie,pai2024survey}. However, because these relations remain grounded in surface lexical forms rather than normalized, domain-specific semantics, they introduce severe heterogeneity and sparsity, sacrificing scalability, domain-specificity, and computational compatibility. As a result, they are rarely used in downstream tasks \cite{niklaus2018survey,romero2023mapping}. 

\textbf{Ambiguity and Coarse Granularity of Relation Labels:} Another fundamental limitation of many existing KGs lies in their insufficient relational granularity for capturing domain-specific distinctions. This issue has been extensively discussed in the biomedical ontology literature, where even seemingly basic relations such as \textit{part-of} exhibit substantial semantic ambiguity. In biomedical contexts, part-of may refer to participation in a biological process, temporal inclusion, or functional dependency, each of which carries different implications. Collapsing these distinct meanings into a single relation type obscures critical distinctions required for accurate reasoning and integration \citep{smith2005relations,schulz2006biomedical}. Similar limitations arise in other domains. For example, relations such as partner and spouse encode different social and legal semantics, yet are often conflated or mapped to a single coarse-grained relation in schema-based KG construction pipelines \citep{biswas2023knowledge}.

\textbf{Context Dependence Beyond Symbolic Relation Granularity:} A natural response to the limitations of coarse relation schemas is to develop more fine-grained symbolic relation types. However, insights from social science suggest that increasing relational granularity alone does not resolve the fundamental challenge of capturing the meaning and context-dependence of real-world relationships. Many real-world relationships are inherently complex, contextual, nuanced, and uncertain, making them difficult to faithfully capture using a single or a few symbolic labels. Conceptual relations, for example, may be disputed, evolving, or dependent on perspective, often requiring a qualitatively descriptive explanation rather than categorical assignment \citep{emirbayer1994network,boccaletti2014structure}. Social relationships provide a particularly clear illustration: Labels such as friendship between two individuals or customership toward a company collapse a wide spectrum of social ties that differ along dimensions such as duration, obligation, intensity, and consumption context. In some cases, a relationship cannot be cleanly categorized, as it may fall within a vague space spanning multiple relationship types. Even attempts to refine such relations using numerical scores or weights remain limited, as scalar values obscure the underlying qualitative distinctions. 

% Add citation in this section
\section{Paradigm Shift and Opportunities in the LLM Era}

The limitations of symbolic relation schemas motivate a reconsideration of how relationships are represented in KGs. Rather than simply refining categorical relation types, we argue for a shift toward representing relations directly in natural language. This shift is enabled by the widespread adoption of LLMs, which prefer contextual text and support downstream reasoning.
At the same time, LLMs have also become central to data and knowledge creation workflows, enabling scalable synthesis and evaluation of relational information expressed in natural language.

\subsection{Natural Language as Input in LLM-Based Downstream Tasks}
 % better to add a couple citations that mention the usage shift from traditional models to LLMs. The focus is that traiditonal model need fix input/output format that is usually not natural language

Prior to the LLM era, information from KGs was typically incorporated into downstream NLP models such as classifiers, rankers, and scoring functions indirectly, through graph-derived features or embeddings rather than being consumed in natural language form. Consequently, KGs needed to be transformed into quantified representations compatible with these models. The most widely used and effective approaches for this transformation include translation-based models such as TransE and its variants \citep{bordes2013translating}, bilinear factorization models such as ComplEx \citep{trouillon2016complex}, as well as graph traversal– or random-walk–based feature generation methods \citep{xia2019random}. These approaches typically assume that relations are drawn from a finite, predefined set of categorical types, which serve to index model parameters, define feature spaces, or guide graph operations. Although recent studies have explored incorporating semantic information from text into KG embeddings or representations \citep{wang2022simkgc,rao2024knowledge,zhu2025text}, such efforts are generally designed to augment structure-based representations derived from categorical relations rather than to replace them as the primary relational abstraction. As a result, relations in KGs are commonly constructed in this form to ensure broad compatibility with downstream models. 
% generic citations regarding LLMs (focus: 1. paradigm shift for model usage; 2. LLM prefers input with nuanced context for better performance)

In contrast, the LLM era has substantially blurred the boundaries between downstream tasks, as many are now approached through in-context learning via prompting \citep{dong2024survey}. Under this paradigm, a single modeling framework can support classification, retrieval, reasoning, or generation, depending primarily on the provided context and instructions. Even tasks that were traditionally formulated as classification with categorical outputs are increasingly reframed as reasoning-oriented problems, in which models take natural language input, generate text-based intermediate reasoning (e.g., chain-of-thought), and ultimately produce textual outputs \citep{wang2024cot, diao2024active}. Empirical evidence from recent work indicates that such reasoning-oriented strategies, particularly chain-of-thought prompting with rich contextual information, often yield better performance than direct prediction \citep{wei2022chain, wang2023towards}. Consequently, LLM-based downstream tasks are no longer tightly coupled to fixed input–output formats and instead favor nuanced, descriptive knowledge provided as context, creating new opportunities for consuming relation knowledge in natural language rather than as simple categorical labels.

% citations about how KG is incorporated into LLM-based downstream tasks. Focus 1. knowledge verbalization - transform relational triples (entity 1, relation type, engity 2)  structures / categorical or numeral features into plain text for better lLM compatibility.
Within the KG literature, this shift has led to growing interest in LLM-based inference that leverages KG knowledge expressed in natural language form \citep{shu2024knowledge, wang2024mgsa, mavromatis2025gnn}. Instead of consuming relational triples in their raw symbolic format or graph-derived structured features, most studies adopt knowledge verbalization strategies that convert KG information into natural language descriptions, which are then incorporated into prompts as contextual input \citep{baek2023knowledge,zhang2024knowgpt}. This KG-to-text conversion is motivated by the observation that LLMs reason more effectively over well-organized natural language representations than over simple symbolic, numerical, or vectorized inputs without verbalization. Consequently, representing relations in natural language within a KG provides a practical path to address the limitations of current representations, while also delivering richer context for downstream LLM reasoning.%Consequently, even when a schema-based KG underlies the system, natural language becomes the primary interface through which relational knowledge is consumed during downstream inference. More detailed relation instead of triples could be  .....

%This shift also has implications for how structured knowledge is consumed by an expanding population of LLM users. As prompt-based interaction with large language models becomes increasingly common, many users who rely on LLMs for downstream tasks do not possess the technical expertise required to incorporate traditional schema-based knowledge graphs into model inputs. Symbolic triples, graph structures, and vectorized representations are not directly compatible with language-based reasoning and typically require nontrivial transformation into textual form before they can be used as contextual input. For such users, the primary challenge lies not in accessing structured knowledge, but in bridging the representational gap between symbolic graph form and natural language. Knowledge graphs with natural-language relations mitigate this gap by providing relational knowledge in a form that can be directly integrated into LLM contexts, thereby enabling the use of structured knowledge without requiring specialized expertise in schema engineering or representation transformation.

\subsection{LLM-Assisted Knowledge Resource Creation}

Recent work increasingly uses LLMs to create new data and knowledge resources \citep{tan2024large,yang2025comprehensive}. Unlike the pre-LLM era, where reusable resources were typically designed around fixed formats and tightly structured representations to satisfy specific downstream models, LLM-based approaches often prioritize descriptive natural language representations that preserve contextual and semantic detail \citep{ding2024data, zheng2025automation}.

% cite papers about generic data/knowledge resource creation using LLMs. Focus 1. the output is from categogical/numeric/short terms etc. to nuanced natural language. Focus 2. the tasks shift from traditional tasks to reasoning, sythesis/summariation, retrieval-generation (not simple retrieval, but extract more concise info - sometimes hidden info)
A wide range of recent studies leverage LLMs to extract, synthesize, or articulate core information from complex documents, such as legal texts \citep{song2025legal,zhou2025lawgpt}
%breton2025leveraging: term extraction is not in the scope hao
and scientific papers \citep{song2025scientific,zheng2025automation}. Examples include claim mining \citep{dmonte2024claim,scire2024fenice}, idea identification \citep{kumar2025can,su2025many}, and knowledge organization and summarization \citep{han2024automating,feng2025enhancing}, where models either extract and synthesize factual information or mine more complex hypothesis, ideas, and arguments. In these settings, the goal is not to only store symbolic labels, short textual terms, or numerical features, but also to create knowledge artifacts that retain nuance in natural language.

% cite papers that using LLM judge to control quality
To enable quality control, many of these approaches combine LLM generation with additional mechanisms. On the generation side, prompts often incorporate explicit rules or constraints to guide content selection and level of abstraction \citep{cheng2024structure,wang2024boosting,mirbeygi2025prompt}. On the evaluation side, LLM-based judges or filtering strategies are increasingly employed to assess relevance, consistency, or factuality, allowing large-scale construction of natural-language knowledge resources with acceptable reliability \citep{liu2024calibrating, li2024llms,huang2025empirical}. Although such resources are less rigidly structured than traditional datasets, they are well aligned with LLM-centered downstream use, where reasoning over textual context is the dominant mode of computation.

% cite how LLM are used in KG construction
Within the KG literature, LLMs have been increasingly explored as tools for KG construction. However, most existing work remains grounded in the traditional KG paradigm, using LLMs primarily to assist with ontology design and refinement, entity and relation extraction, or knowledge completion via link prediction \citep{wei2023kicgpt,fathallah2024neon,liang2024survey,zhu2024llms}. In these approaches, LLMs largely function as extractors or classifiers, while the resulting KGs continue to rely on predefined symbolic relation schemas. A small number of studies have begun to enrich entity representations with more nuanced natural-language descriptions \citep{zhu2025context}, yet analogous advances for relation representation remain largely unexplored.

\section{Principles for Structure–Expressiveness Trade-offs}

Despite the expanded flexibility introduced by LLMs, fundamental challenges in leveraging KGs remain. In particular, LLMs are unable to ingest an entire KG within their context window, both due to input token limits and because excessively long contexts incur higher computational cost and may degrade performance \citep{du2025context,li2025longcontext}. As a result, Retrieval-Augmented Generation (RAG) is still the most common way to use LLM \citep{arslan2024survey}, which first requires to identify and retrieve the most relevant subsets of knowledge for a given downstream task. Consequently, effective KG utilization in the LLM era still relies on mechanisms for efficient information retrieval, often supported by well-defined schemas and graph-based algorithms or embeddings that commonly take a fixed set of symbolic relations as graph edges.

This requirement highlights a persistent tension among \textbf{expressiveness}, referring to the level of detail in relational descriptions, \textbf{generalizability}, referring to the range of domains or tasks to which a KG can be adapted, and \textbf{structural regularity}, which typically relies on abstracted symbolic relation types. While natural language relations allow richer and more nuanced representations of relational knowledge, purely descriptive relations substantially complicate graph traversal, indexing, and retrieval. By contrast, categorical or symbolic relations continue to provide an efficient and effective means of organizing global graph structure and supporting path-based reasoning and relevance filtering.

These challenges suggest that the LLM era does not eliminate the need for structure in KGs, but instead reshapes how structure and language should interact within a knowledge representation framework. Rather than replacing symbolic relations entirely, natural-language relations must be designed in a way that preserves structural usability while enabling richer semantic expression.

Motivated by this trade-off, we outline a set of guiding principles for constructing natural-language-relation–based KGs that balance structural tractability with linguistic flexibility.

\subsection{Implicit, Task-Aware Schemas Remain Necessary}

Although natural-language relations relax the rigidity imposed by fixed symbolic schemas, some form of schema remains essential for effective use. Rather than being fully open-ended, relation generation and representation should be guided by implicit, task-aware schemas derived from clearly defined goals, domains, and downstream requirements. In particular, relations should be constrained to an appropriate semantic scope: introducing a large number of heterogeneous or cross-domain relation types often increases ambiguity and noise, ultimately hindering retrieval and reasoning rather than improving expressiveness. At the same time, schemas should not be overly explicit or rigid, as fixed ontologies can limit generalization across tasks and domains.

In practice, effective relation synthesis, especially when integrating knowledge from multiple sources, requires an explicit specification of the intended goal, scope, and domain of the resulting relations. When prompting LLMs to synthesize knowledge materials into concise relational descriptions, constraints or guiding rules are often necessary to control granularity, scope, and relevance. Such implicit, scoped schemas provide structural guidance without enforcing a rigid ontology, thereby balancing flexibility, generalization, and practical usability.

\subsection{Natural-Language Relations Should Support Secondary Symbolic Inference}

Closely related to the need for implicit, task-aware schemas, natural-language relation descriptions should not be treated as isolated language artifacts. Instead, they should be designed to support secondary inference into symbolic relations when required. In downstream settings that rely on symbolic representations, such as structured querying, aggregation, or graph-based retrieval, natural-language relations should be readily classifiable or mappable to existing symbolic categories.

From this perspective, natural-language relations are not intended to replace symbolic relations with an incompatible representation, but rather to serve as a richer carrier of relational information that remains compatible with symbolic abstraction when needed. This compatibility enables natural-language relations to preserve fine-grained semantic content while still supporting structure-dependent operations, thereby bridging flexible language-based representation and traditional symbolic reasoning.

\subsection{Hybrid Use of Symbolic Relations and Natural-Language Descriptions for KG Construction}

A hybrid design that combines a small set of symbolic relations with natural-language relation descriptions provides a practical compromise between structural stability and representational flexibility. Symbolic relations establish an initial global backbone for the graph, enabling efficient organization, traversal, and retrieval, while natural-language descriptions capture nuanced, contextual, or evolving aspects of relationships that are difficult to encode in fixed categorical form. For example, two entities may be connected by a symbolic friend relation, accompanied by a natural-language description that specifies the nature and context.

In this hybrid setting, the symbolic relation schema is best designed to be generic, domain-agnostic, and minimally committal, serving primarily as a guiding structural scaffold rather than a complete semantic specification. Such symbolic relations provide coarse-grained relational anchors that support graph-level operations, while leaving fine-grained semantics to be expressed in natural language. Importantly, because rich relational meaning is preserved in textual form, symbolic relations can be more easily revised, refined, or restructured, such as being split, merged, or redefined, based on insights derived from natural-language descriptions. This design enables the KG to evolve without requiring disruptive changes to its global structure, supporting both scalability and long-term adaptability. This hybrid design directly reflects the first two principles: symbolic relations provide an implicit, task-aware structural schema, while natural-language descriptions preserve sufficient semantic detail to enable secondary inference into symbolic relations when needed.

\section{Future Research Directions}

While the principles outlined above provide high-level guidance for integrating natural-language relations into KGs, translating these principles into practical systems raises a number of open challenges and realistic considerations. Constructing and maintaining KGs that balance structural tractability, domain and task generalizability, and contextual nuance requires coordinated progress across data acquisition, representation, retrieval, inference, and evaluation. In this section, we outline several promising directions for future research aimed at overcoming these challenges and advancing the development of KGs grounded in natural-language relations.

\subsection{Constructing Natural-Language Relation KGs under Heterogeneous Sources}

A fundamental challenge in constructing natural-language-relation KGs from raw text is the prevalence of heterogeneous and conflicting information across sources. Classical computer science literature typically addresses this problem under the frameworks of data fusion, knowledge fusion, or truth discovery, where conflicting facts are reconciled by estimating source reliability or latent truth, often with the goal of selecting the most ``correct'' value rather than preserving disagreement \citep{wang2024survey}. While effective for many objective settings, such approaches are less suitable for domains such as history or social narratives, where conflicting accounts may all be plausible and should be retained rather than forcibly resolved. In the context of natural-language-relation KGs, a key research question is how to synthesize and preserve conflicting relational descriptions in a concise yet faithful manner \citep{info15080509}, leveraging LLMs’ reasoning capabilities without collapsing uncertainty. 

\subsection{LLM-Driven Refinement of Existing KGs}
Beyond constructing KGs from raw text, an alternative and complementary direction is to start from existing KGs and leverage additional data sources, or even the internal knowledge and reasoning capabilities of LLMs, to enrich symbolic relations with more nuanced natural-language descriptions. For example, recent studies have explored distilling relational knowledge from LLMs in natural language for KG completion, while still operating within a symbolic relation paradigm and targeting symbolic relation prediction \citep{li2024contextualization,yao2025exploring}.

However, enriching existing KGs with natural-language relations introduces additional challenges when the underlying symbolic schema itself is limited or misaligned with the newly added semantic content. In such cases, KG modification cannot be restricted to appending information on top of fixed relation types, but may require simultaneous refinement of the relation schema, including revising, splitting, or merging symbolic relations based on richer relational evidence. This raises open research questions about how to coordinate LLM-driven semantic enrichment with controlled schema evolution, while preserving structural consistency and downstream usability.

\subsection{Contextual Relation–Based Traversal and Reasoning}

Most existing KG traversal methods rely primarily on graph structure, such as neighborhood expansion, path-based reasoning, or graph embedding similarity, treating relations as fixed symbolic labels or parameters \citep{zhu2025text}. In natural-language-relation KGs, however, the reasoning capabilities of LLMs enable traversal decisions to be made dynamically based on the semantic content of relations and the evolving context of the task. Rather than treating traversal as a fixed graph operation, LLMs can iteratively select relevant nodes, relations, or subgraphs by reasoning over natural-language descriptions at each step. This opens the possibility of multi-stage reasoning pipelines, where LLMs are repeatedly invoked to guide retrieval, assess relevance, and refine the search space over the KG.

\subsection{KG Embeddings}
Another promising research direction is the development of joint structural–semantic embedding methods for KGs with natural-language relations. Prior work has incorporated semantic information into KG embeddings by leveraging relation names, textual descriptions of entities, or auxiliary text encoders, demonstrating improvements over purely symbolic representations \citep{wang2017knowledge,yao2019kg,cao2024knowledge}. However, the relation semantics considered in these approaches are typically simple, static, and limited in scope, and are largely treated as auxiliary signals rather than primary relational representations. In contrast, the hybrid design advocated in Principle 3, where a small set of generic symbolic relations provides global structure and natural-language descriptions, opens new opportunities for embedding models that jointly capture structural regularities and nuanced semantic content.

\subsection{KG Evaluation}
Evaluating KGs with natural-language relations presents a significant challenge because relations are no longer fixed symbolic types or simple lexical surfaces, but synthesized descriptions produced by LLMs. As a result, traditional KG metrics such as precision and recall over triples are insufficient, as they fail to capture key qualities of natural-language relations like factual faithfulness, core information coverage, and conciseness. Recent work has therefore explored LLM-as-a-judge frameworks that assess generated text using rubric-based prompting \citep{liu2023g, es2024ragas,li2025generation}. Adapting such judge-based approaches to KG evaluation introduces new questions around consistency, bias, and scalability, motivating evaluation protocols tailored to natural-language-relation KGs.

\section{Conclusion}
KGs have long relied on symbolic relation schemas to enable scalable construction and broad computational use. The rise of LLMs reshapes this foundation by making natural language the primary interface for downstream inference. This paper argues that treating relations as natural-language descriptions, supported by implicit and hybrid structural constraints, provides a practical way to align KGs with current LLM usage paradigms while preserving nuanced relational information. Rather than viewing LLMs only as more efficient tools for populating conventional schemas, we advocate rethinking the relations themselves as contextual language descriptions that remain supported by a minimal symbolic backbone for retrieval and traversal. Building on this perspective, we further advocate the development of new methods and empirical research practices under this paradigm.

\section{Limitations}
This position paper has three main limitations.

First, we focus primarily on the paradigm shift for relations in KGs, while leaving analogous changes for entities less examined. In the LLM era, entity extraction and presentation also require substantial reconsideration. By addressing only the relational side, our discussion captures a partial view of the broader KG transformation. Future research should jointly rethink entity and relation representations so that both components align with prompting-based inference and LLM-centered knowledge usage.

Second, our arguments are developed largely around textual KGs constructed from language corpora. KGs may also be grounded in images, audio, or other modalities, yet this work does not explore such multi-modal settings. Relations expressed across heterogeneous modalities introduce additional challenges that differ from purely text-based scenarios. Consequently, the conclusions here should be interpreted within the limits of language-centered KG construction rather than as a general account of all KG forms.

Finally, this is a position paper that provides conceptual analysis based on a synthesis of prior literature, without presenting new large-scale experiments. The discussion therefore remains high-level and partly theoretical, and practical effectiveness of natural-language relations is not empirically validated here. The proposed principles serve as guidance rather than tested protocols, and implementing them in real systems will require dedicated methodological research and systematic evaluation.

\bibliography{custom}

\end{document}